\documentclass{article} %
\usepackage{arxiv,times}

\usepackage{amsmath,amsfonts,bm}

\def\eqref#1{equation~\ref{#1}}

\def\1{\bm{1}}

\DeclareMathAlphabet{\mathsfit}{\encodingdefault}{\sfdefault}{m}{sl}
\SetMathAlphabet{\mathsfit}{bold}{\encodingdefault}{\sfdefault}{bx}{n}

\usepackage[hidelinks]{hyperref}
\usepackage{url}
\usepackage{booktabs}
\usepackage{multirow}
\usepackage{graphicx}
\usepackage{makecell}
\usepackage[table]{xcolor}
\usepackage{enumitem}

\definecolor{improvegreen}{HTML}{D4EFD4}
\definecolor{mathbg}{HTML}{FCEEE8}   %
\definecolor{codebg}{HTML}{ECECFA}   %
\definecolor{genbg}{HTML}{E8F1FA}    %
\definecolor{avgbg}{HTML}{F2F2F2}

\newcommand{\mockreduce}[1]{%
  \cellcolor{improvegreen}\ensuremath{\downarrow\,#1\%}}
\newcommand{\up}[1]{%
  \ensuremath{\uparrow\,#1}}
\newcounter{algorithm}
\renewcommand{\thealgorithm}{\arabic{algorithm}}
  
\title{Latent Bridge: Asking for Help Without Words}

\author{%
Qichao Wang\textsuperscript{1}\thanks{Equal contribution.},
Tingyu Qu\textsuperscript{2,*},
Pengcheng Wu\textsuperscript{1,*},
Yiran Zhong\textsuperscript{2} \\
Feida Zhu\textsuperscript{2},
Zhiqi Shen\textsuperscript{1}\thanks{Corresponding authors.},
Peilin Zhao\textsuperscript{3,\textdagger},
Steven Hoi\textsuperscript{2} \\[0.55em]
{\normalfont\small \textsuperscript{1}Nanyang Technological University} \\
{\normalfont\small \textsuperscript{2}Alibaba Token Hub, Alibaba Group} \\
{\normalfont\small \textsuperscript{3}Shanghai Jiao Tong University}%
}

\hypersetup{
  pdftitle={Latent Bridge: Asking for Help Without Words},
  pdfauthor={Qichao Wang, Tingyu Qu, Pengcheng Wu, Yiran Zhong, Feida Zhu, Zhiqi Shen, Peilin Zhao, and Steven Hoi}
}

\usepackage{xspace}
\newcommand{\method}{\textsc{POND}\xspace}

\arxivfinalcopy
\begin{document}

\maketitle

\begin{abstract}
Text-based model collaboration requires a large model to generate intermediate responses before a smaller model can use its guidance, adding large-model decoding cost to inference. We introduce \method{} (\textbf{P}refill-\textbf{O}nly, \textbf{N}o-\textbf{D}ecode), a method that uses a frozen large sender for prefill and a small receiver for all autoregressive decoding. A trainable bridge maps the sender's hidden states to additional attention keys and values, allowing the receiver to use these representations throughout generation. \method{} also supports periodic review, in which the sender processes the receiver's partial response and updates its guidance through additional prefill passes while preserving the receiver's existing KV cache. Across reasoning and coding benchmarks, \method{} improves receiver performance with senders from different model families. In particular, GLM-5.3 provides gains comparable to those of Qwen3.8-Max when paired with a Qwen receiver, supporting the effectiveness of representation transfer between independently trained model families. 
\method{} further demonstrates its ability to generalize across domains. When trained on math and coding tasks, it transfers directly to general knowledge tasks without additional domain-specific training. These results demonstrate that large frontier model prefill representations can improve small-model generation without requiring large-model autoregressive decoding.
\end{abstract}

\section{Introduction}

Natural language is the primary means of human communication, and multi-model systems often mimic this approach by exchanging generated text. 
One model generates an intermediate response that another model then uses as input~\citep{du2023improving,wu2023autogen,wang2406mixture}. Text supports communication between models with different architectures, 
but requires the sender to generate an intermediate response and then the receiver to process it. 
Such communication can be particularly costly when a frontier-scale model assists a smaller model deployed on edge hardware.
In this setting, the large model decodes the intermediate response that will be sent the small model token by token, so the system incurs large-model decoding latency even though the small model produces the final answer.
These limitations motivate a key question:

\vspace{-0.3em}
\begin{center}
\textit{Can a large model improve a small model’s generation without incurring the latency of autoregressive decoding?}
\end{center}
\vspace{-0.3em}

Our approach rests on an asymmetry in large-model inference cost. During prefill, a model processes all input tokens in parallel; the computation is compute-bound and uses the hardware efficiently. Standard autoregressive decoding instead requires one forward pass per generated token. 
These passes must run sequentially, so total decoding latency grows with response length, making long responses particularly time-consuming to generate~\citep{patel2024splitwise, zhong2024distserve}.
Prefill produces contextual representations of the input without requiring the large model to generate text. We therefore investigate how much of the benefit of large-model assistance can be recovered by transferring these representations directly to a small model.

We refer to the large model as the \emph{sender} and the small model as the \emph{receiver}. The main challenge is that the two models are trained independently. Their hidden dimensions differ, and even at matched dimensions their representation spaces are not aligned, so the receiver cannot use the sender's states directly. We address this with a trainable bridge that maps sender hidden states to additional attention keys and values in the receiver. Both models read the same input, and the receiver attends to the bridged keys and values alongside its own context. The bridge is trained with the receiver's next-token prediction loss on target answers, while the sender remains frozen.

We introduce \method{} (Prefill-Only, No-Decode), 
a framework for
collaboration between a frontier-scale sender and a small receiver
capable of running on suitable edge hardware.
The sender performs only prefill, while the receiver handles all autoregressive decoding. 
Throughout generation, the receiver attends to keys and values
derived from the sender's representations while retaining its
original prompt and KV cache.
This design allows the receiver to 
leverage frontier-model representations 
without updating the sender's parameters or requiring it to generate intermediate tokens.

A related approach is multi-token prediction (MTP), which trains additional prediction heads or modules to predict multiple future tokens from a model's hidden states~\citep{gloeckle2024better,liu2024deepseek}. These modules can also be used for speculative decoding, where the main model verifies proposed tokens. \method{} instead transfers hidden states between separately pretrained models and trains the bridge using the receiver's next-token prediction objective. The receiver generates the final response directly, without sender verification or a requirement to preserve the sender's output distribution. The goal is to improve small-model generation using large-model prefill, rather than to accelerate sampling from the large model.

We also extend \method{} with \emph{periodic review}. In the initial-only setting, the sender processes the prompt once. With review, the sender periodically processes the available response prefix through an additional prefill pass. 
The bridge maps updated sender representations to keys and values, while the receiver continues decoding from its existing KV cache.
Review allows sender guidance to depend on the partial response while keeping all autoregressive decoding in the receiver.

We evaluate POND on math, code, and general-knowledge tasks using three frontier-scale senders and an edge-scale receiver. Initial transfer alone improves receiver performance. Periodic review provides further gains, particularly for tasks requiring long-form generation. By updating the sender representations based on the partial response, periodic review maintains effective guidance over extended decoding. These results show that frontier-model prefill representations can effectively guide small-model generation without requiring large-model autoregressive decoding. Periodically updating these representations further improves their effectiveness for long outputs.

Our contributions are:
\begin{itemize}[leftmargin=1.2em, itemsep=2pt, topsep=2pt, parsep=0pt]
\item We propose \method{}, a collaboration framework in which a frozen large sender performs only prefill and a small receiver performs all autoregressive decoding. A trainable bridge maps sender hidden states to additional attention keys and values in the receiver, allowing the two independently trained models to communicate without intermediate text.
\item We introduce periodic review, which lets the sender re-read the partial response through additional prefill passes. 
This allows guidance to adapt to the subsequent response without requiring sender-side decoding or recomputation of the receiver’s generated prefix.
\item We show that, across frontier-scale senders, \method{} consistently improves receiver performance on math and code tasks and transfers directly to general knowledge tasks without additional training. These results highlight \method{}’s cross-domain generalizability.
\end{itemize}

\section{Related Work}
\subsection{Model Communication and Collaboration}

Many multi-model systems improve performance by allowing language models to exchange natural-language messages. 
Prior work has explored communication both within a single
model~\citep{qu2024routing,sun2026event,deepseek2026v41flash}
and across multiple models. Multi-agent debate, for example,
lets several models propose, critique, and revise their solutions over multiple rounds~\citep{du2023improving,chen2024reconcile}.
General agent frameworks organize models into specialized roles and coordinate their work through conversation \citep{wu2023autogen,qian2024chatdev}. Mixture-of-Agents follows a layered design in which each model conditions on the textual outputs produced by models in the previous layer \citep{wang2406mixture}. Other work studies how the communication topology can be made sparse to reduce the token and inference cost of multi-agent discussion \citep{li2024improving}.

Text offers an interpretable interface across heterogeneous models, but incurs sender-side decoding costs and limits communication to what the sender expresses in text.
Beyond text-based communication, Cache-to-Cache~\citep{fu2026cachetocache} enables direct semantic transfer by projecting and fusing KV caches across models.
LLM-to-SLM~\citep{bergner2024think} uses a frozen large language model to encode the prompt once in parallel and conditions a smaller model on the resulting representations to generate the response.
\method learns a latent channel that transforms frontier-model hidden states into a separate guidance memory, preserving the receiver's text KV cache.
This interface supports both initial transfer and response-dependent periodic review, allowing the sender to provide guidance entirely through prefill while the receiver handles all autoregressive decoding.

\subsection{Knowledge Distillation and Latent Reasoning}

Knowledge distillation trains a smaller student to reproduce information provided by a larger teacher, commonly through output distributions or teacher-generated sequences \citep{hinton2015distilling,kim2016sequence}. Feature-based methods additionally match hidden states, attention patterns, or other intermediate representations \citep{jiao2020tinybert,wang2020minilm}. For generative language models, recent methods distill teacher distributions, supervise the student on its own generated sequences, or use teacher-generated rationales as additional training targets \citep{gu2024minillm,agarwal2024policy,hsieh2023distilling}. 
These methods typically transfer teacher knowledge into a standalone student that operates without the teacher at inference time.

\method instead trains a latent bridge and selected receiver components
to use guidance from a frozen sender at inference time. This guidance
depends on the current prompt and, under periodic review, on the
receiver's partial response. Rather than aiming to produce a standalone
student, POND transfers instance-specific computation through prefill
representations, without requiring sender-side autoregressive decoding.

A related line of work explores intermediate computation beyond the final visible output. 
Pause tokens and internal rationales enable additional computation before prediction~\citep{goyal2024think,zelikman2024quiet}, while latent-reasoning methods use hidden-state feedback or compressed
chain-of-thought representations~\citep{hao2024training,shen2025codi}.
These methods primarily study reasoning within a single model, whereas \method{} transfers prefill representations between independently trained models without sender-side autoregressive decoding.

\section{Method}
\label{sec:method}

\method connects a large sender model to a small receiver
through a learned latent bridge. The sender only reads, through prefill passes; the receiver generates
every output token.
A lightweight bridge turns the sender's hidden
states into a latent message that the receiver reads as additional
KV entries.

\begin{figure}[ht]
    \centering
    \includegraphics[width=\linewidth]{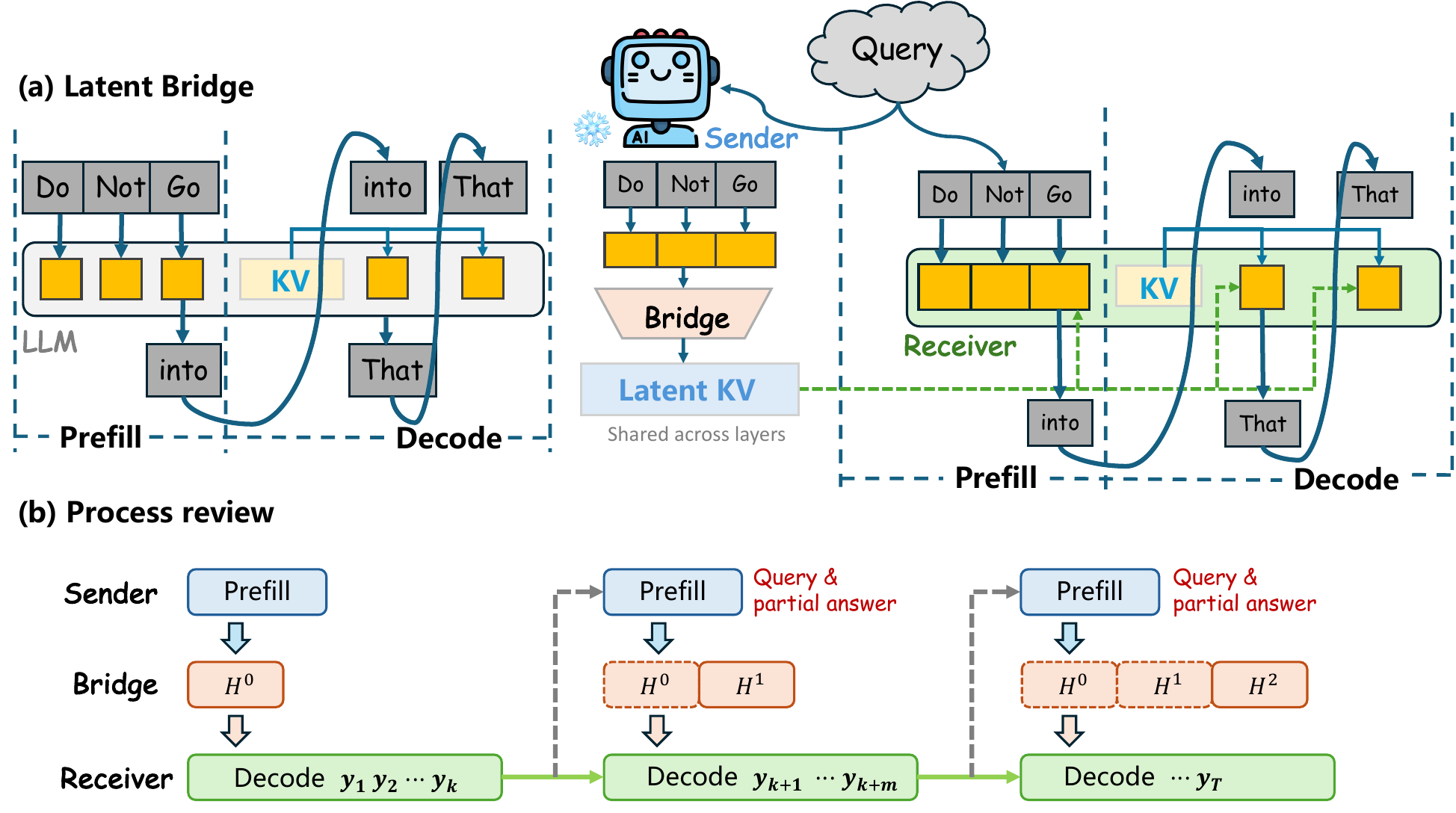}
    \caption{Overview of \method. A frozen sender prefills the input,
    and a trainable bridge compresses its last layer hidden states
    into a shared latent message, which layer-specific adapters project
    into KV entries for the receiver. The receiver, fine-tuned jointly
    with the bridge, generates the full response. At review boundaries,
    the sender prefills the newly generated segment and the resulting
    entries are appended to the latent memory, leaving the receiver's
    KV cache untouched.}
    \label{fig:architectur}
\end{figure}

\subsection{Problem Setup}
\label{sec:overview}
 
Let $x$ be the input and $y=(y_1,\ldots,y_N)$ the output. We denote
the frozen sender by $S_{\theta_s}$, the receiver by $R_{\theta_r}$,
and the bridge parameters by $\phi$. In the receiver's token space,
$x$ occupies positions $1,\ldots,n_x$ and $y_t$ occupies position
$n_x+t$. The receiver generates
\begin{equation}
    p_{\theta_r,\phi}(y \mid x)
    =
    \prod_{t=1}^{N}
    p_{\theta_r,\phi}
    \bigl(y_t \mid x,\, y_{<t},\, \mathcal{M}_t\bigr),
    \label{eq:conditional_generation}
\end{equation}
where $\mathcal{M}_t$ is the latent memory available at step $t$:
a single message computed from $x$ in the basic setting, enlarged
during generation by periodic review.
 
\subsection{Latent Compression}
\label{sec:compression}

We take the final-layer hidden states of the sender,
$H = S_{\theta_s}(x) \in \mathbb{R}^{T\times d_S}$, where $T$ is the
sender input length and $h_j$ denotes the $j$-th row.
The bridge maps each state to a low-dimensional latent,
\begin{equation}
    z_j
    =
    \operatorname{MLP}_{\phi}(h_j) \in \mathbb{R}^{d_z},
    \qquad d_z \ll d_S,
    \label{eq:compression}
\end{equation}
giving the message $Z = B_\phi(x) = [z_1;\ldots;z_T]$.
Compression acts on the feature dimension only: each sender token
keeps one latent, so every entry has a well-defined location in the
text, and a single $Z$ is shared by all receiver layers. Because the
sender is causal and $B_\phi$ acts position-wise, each latent depends
only on its own prefix,
\begin{equation}
    z_j = f_\phi\bigl(x_{\le j}\bigr),
    \label{eq:prefix_determinism}
\end{equation}
a property we call \emph{prefix determinism}. It enables both
incremental review and single-pass training.

\subsection{Layer-Specific KV Adapters}
\label{sec:kv_adapters}
 
We attach an adapter to each full-attention layer of the receiver.
For an adapted layer $\ell$, a module $\mathcal{P}_{\phi_\ell}$ (K/V
projections followed by RMS normalization) maps the shared latent to
keys and values in the receiver's KV-head layout,
\begin{equation}
    \bigl(\bar{K}_{\ell}^{z}, V_{\ell}^{z}\bigr)
    =
    \mathcal{P}_{\phi_\ell}(Z)
    \in
    \mathbb{R}^{T \times n_{\mathrm{kv}} \times d_h},
    \qquad
    K_{\ell,j}^{z}
    =
    \operatorname{RoPE}_{\ell}
    \bigl(\bar{K}_{\ell,j}^{z},\, p_j\bigr),
    \label{eq:kv_projection}
\end{equation}
where $p_j$ is the receiver position aligned with latent $j$. The entries are computed once per
message and cached. Sharing $Z$ keeps the message compact, while
per-layer adapters let each layer read it in its own subspace.
 
\subsection{Gated Joint Attention}
\label{sec:gated_attention}
 
At an adapted layer $\ell$, let $Q_\ell$ be the receiver queries and
$K_\ell^{\mathrm{txt}},V_\ell^{\mathrm{txt}}$ the keys and values of
the input and the available output prefix, under the usual causal
mask. Alongside the original output
$A_\ell^{\mathrm{txt}} = \operatorname{Attn}(Q_\ell,
K_\ell^{\mathrm{txt}}, V_\ell^{\mathrm{txt}})$, the receiver computes a
joint output normalized over the visible latent entries and the text,
\begin{equation}
    A_\ell^{\mathrm{joint}}
    =
    \operatorname{Attn}
    \bigl(
        Q_\ell,
        [K_\ell^{z};K_\ell^{\mathrm{txt}}],
        [V_\ell^{z};V_\ell^{\mathrm{txt}}]
    \bigr),
    \label{eq:joint_attention}
\end{equation}
and combines the two through a learned gate $g_\ell$ initialized to
zero:
\begin{equation}
    A_\ell
    =
    A_\ell^{\mathrm{txt}}
    +
    \tanh(g_\ell)
    \bigl(A_\ell^{\mathrm{joint}} - A_\ell^{\mathrm{txt}}\bigr).
    \label{eq:gated_joint_attention}
\end{equation}
 
\paragraph{A residual view.}
Let $A_\ell^{z}$ be attention over visible latent entries alone,
and $\lambda_\ell^{z}$, $\lambda_\ell^{\mathrm{txt}}$ the log-sum-exp
of the latent and text logits. The joint softmax factorizes with
$\alpha_\ell = \sigma(\lambda_\ell^{z} - \lambda_\ell^{\mathrm{txt}})$,
so that
\begin{equation}
    A_\ell
    =
    A_\ell^{\mathrm{txt}}
    +
    \tanh(g_\ell)\,\alpha_\ell
    \bigl(A_\ell^{z} - A_\ell^{\mathrm{txt}}\bigr).
    \label{eq:residual_view}
\end{equation}
The latent branch is thus a residual correction scaled by a static
gate and a content-dependent coefficient $\alpha_\ell$, and it
vanishes whenever no latent entry is visible. The identity also gives
an efficient implementation: text attention runs through the
receiver's unmodified kernel, and a small latent attention is merged
through log-sum-exp statistics, as in split-KV attention.
 
\subsection{Periodic Review}
\label{sec:periodic_review}
 
Let $\mathcal{B} = \{b_r \mid r = 0,\ldots,R,\;
0 = b_0 < b_1 < \cdots < b_R < N\}$ denote the review boundaries.
At each $b_r \in \mathcal{B}$,
the sender reads the
partial response $c_r = x \oplus y_{1:b_r}$.
By prefix determinism, only the new segment $y_{b_{r-1}+1:b_r}$ produces new latents. The review message is $Z^{(r)} = [z_j]_{j\in J_r}$, where $J_r$ indexes the segment's sender tokens ($Z^{(0)} = B_\phi(x)$). The sender computes it by prefilling only this segment using its
cached prefix. 
The projected entries are appended to the latent memory,
\begin{equation}
    \mathcal{M}_{\ell}^{(r)}
    =
    \mathcal{M}_{\ell}^{(r-1)}
    \oplus
    \bigl(K_{\ell}^{z,(r)},V_{\ell}^{z,(r)}\bigr),
    \label{eq:memory_update}
\end{equation}
while the receiver's KV cache is left untouched. Since each token
passes through the sender once, total sender computation equals a
single prefill over $x\oplus y_{1:b_R}$ regardless of the number of
reviews; review frequency trades  freshness of guidance against the
number of sender calls and  memory size.
 
\subsection{Causal Visibility and Position Alignment}
\label{sec:visibility}
 
\paragraph{Position alignment.}
With a shared tokenizer, $p_j$ is the receiver position of 
token $j$.
Otherwise we align by character offsets: with $e_j$ and $E_q$
the end offsets of sender token $j$ and receiver token $q$,
\begin{equation}
    p_j = \min\,\{\, q : E_q \ge e_j \,\},
    \label{eq:alignment}
\end{equation}
the first receiver position that has read all text encoded by $z_j$.
Offsets are represented in a shared UTF-8 byte coordinate system; tokens
without an unambiguous receiver counterpart are excluded from latent memory.
Appendix~\ref{app:alignment} gives the full alignment and availability rules.
 
\paragraph{Visibility mask.}
Review $r$ is issued after $y_{b_r}$ is sampled but before the receiver
processes it, 
so its release position is $\rho_r = n_x + b_r$, with $\rho_0 = 0$. 
A query at position $q$ attends to entry $j$ of message
$r$ if and only if
\begin{equation}
    \tau_j = \max(p_j, \rho_r) \le q.
    \label{eq:visibility}
\end{equation}
Prompt positions thus see the initial message causally by position,
response positions see all of it, and review messages become visible
upon release. Every visible entry encodes only text already read at
$q$, and the mask reproduces exactly the entries available at
inference, so parallel training on full sequences matches incremental
deployment.
 
\subsection{Sparse Latent Memory}
\label{sec:sparse_memory}
 
As the memory grows with every reviewed token, we read it with sparse
attention, following Compressed Sparse Attention~\citep{deepseekv4}, which builds on DeepSeek Sparse
Attention~\citep{deepseekv32}; the receiver's text attention is
unchanged. Latent entries are compressed along the sequence at ratio
$m$, and a lightweight indexer keeps the top $k$ visible entries per query,
\begin{equation}
    \mathcal{I}_\ell(q)
    =
    \operatorname{TopK}_k
    \bigl\{\, s_\ell(q, j) \,:\, \tau_j \le q \,\bigr\},
    \label{eq:topk}
\end{equation}
over which $A_\ell^{z}$ and $\lambda_\ell^{z}$ are computed.
Compression blocks do not span messages, a compressed entry takes the
largest threshold among its constituents, and masking precedes
selection, so the guarantees of Section~\ref{sec:visibility} are
preserved.
After candidate selection, latent value aggregation costs
$\mathcal{O}(k)$ per query. Exact query--key scoring can still touch every
visible latent key, so top-$k$ selection does not make total runtime
independent of memory size. The main experiments use the dense configuration
($m=1$); top-$k$ reading is evaluated only as an optional deployment variant.
Appendix~\ref{app:sparse} specifies its masking, tie-breaking, and approximation
properties.
 
\subsection{Training}
\label{sec:training}
 
With the sender frozen, we train the bridge $\phi$ (compressor,
adapters, gates, and indexer) and all receiver parameters $\theta_r$
by full-parameter fine-tuning:
\begin{equation}
    \min_{\phi,\,\theta_r}\;
    -\sum_{(x,y)\in\mathcal{D}}
    \sum_{t=1}^{N}
    \log p_{\theta_r,\phi}\bigl(y_t \mid x, y_{<t}, \mathcal{M}_t\bigr),
    \label{eq:training}
\end{equation}
where $\mathcal{M}_t$ follows the mask of Equation~\ref{eq:visibility}
and review boundaries are sampled per example from the inference
schedule.
Under teacher forcing, prefix determinism makes every review message a
slice of the bridge output on $x\oplus y$, so a single sender prefill
yields all messages and reviews enter training only through the mask.
Zero-initialized gates make the adapted receiver identical to the
original at the start of training. Full-parameter tuning of the
receiver consistently outperforms training the bridge alone.

\section{Experiments}
\label{sec:experiments}

We evaluate \method in two settings: a single prefill-only transfer, where the sender reads the input once, and periodic review, where the sender additionally reads the receiver's partial response at selected boundaries.

\subsection{Experimental Setup}
\label{sec:setup}

\paragraph{Models and comparison conditions.}
Qwen3.5-4B~\citep{qwen35blog} is the receiver in all conditions. The senders are Qwen3.8-Max~\citep{qwen38}, GLM-5.3~\citep{glm5team2026glm5vibecodingagentic}, and DeepSeek-V4-Flash~\citep{deepseekai2026deepseekv4}, providing one within-family and two cross-family pairings.
The sender remains frozen, while the receiver and bridge are jointly optimized using the conditional likelihood in Equation~\ref{eq:conditional_generation}.
\emph{Receiver alone} generates from the original prompt without the bridge.
\emph{\method{}} supplies latent KV entries from one sender prefill pass and makes no further sender calls.
\emph{+Review} additionally processes the available response prefix and appends latent updates as described in Section~\ref{sec:periodic_review}.
In both bridged conditions, all output tokens are generated by the receiver.
The receiver-alone comparison measures the combined effect of receiver adaptation, bridge training, and sender guidance, so it does not isolate the contribution of sender information from that of receiver training. The \method{}--Review comparison measures the additional benefit and cost of periodic sender updates under matched training and evaluation protocols.

\paragraph{Tasks and quality metrics.}
We report AIME (2022-2026)~\citep{maa_aime} and OlymMath-Hard~\citep{sun-etal-2026-challenging} for mathematics; HumanEval+, MBPP+~\citep{NEURIPS2023_humaneval_plus, evalplus2023}, and LiveCodeBench v6~\citep{jain2025livecodebench} for code; and GPQA-Diamond~\citep{rein2024gpqa}, MMLU-Redux~\citep{gema-etal-2025-done}, MMLU-Pro~\citep{wang2024mmlupro}, and ARC-Challenge~\citep{clark2018thinksolvedquestionanswering} for knowledge-intensive question answering.
Acc. denotes exact-match accuracy for math and multiple-choice tasks and pass@1 for code, in percent.
Avg is the unweighted mean of task scores within each suite, not accuracy pooled over examples.

\paragraph{Efficiency metrics and aggregation.}
Token reports decoded output in thousands of tokens per 100 runs; an entry of $v$ corresponds to $10v$ tokens per run.
It excludes sender prefill tokens and latent KV memory and is not a measure of total FLOPs.
Latency reports end-to-end time in seconds per run, including sender prefill, bridge computation, and receiver decoding; lower is better.
Suite-level token and latency reductions are calculated from the unweighted per-task means as $100(1-m/m_{\mathrm{receiver}})$.
This ratio of means differs from the mean of task-wise percentage reductions and gives greater influence to tasks with larger baseline costs.
Accuracy gains are absolute percentage-point differences, calculated before rounding.
The Improve columns select the better bridged value separately for each metric; the joint comparisons below instead use explicitly named configurations.

\subsection{Main Results on Math and Code}
\label{sec:main}

\paragraph{Single-transfer guidance accounts for most of the accuracy gain.}
Table~\ref{tab:main} reports higher accuracy than the receiver for all 15 sender--task pairs even without review.
From the receiver's 63.94\% macro-average, single-transfer \method{} reaches 67.14\%, 66.74\%, and 66.2\% with Qwen3.8-Max, GLM-5.3, and DeepSeek-V4-Flash, respectively.
Review raises these to 68.02\%, 67.42\%, and 66.9\%, improvements of 4.08, 3.48, and 2.96 points over the receiver.
Single-transfer \method{} accounts for 
roughly 80\%
of the final gain, so most of the improvement does not require repeated sender calls.

\paragraph{Task-level gains are heterogeneous.}
Table~\ref{tab:main} shows the largest +Review gains on AIME (5.1--6.2 points) and LiveCodeBench (4.2--5.0), compared with 1.9--2.6 on HumanEval+ and 2.0--2.9 on MBPP+.
OlymMath-Hard improves by 2.8--3.7 points despite a baseline of only 50.0\%, less than the gain on AIME with its 80.0\% baseline.
Baseline headroom alone therefore does not explain the variation.
Overall, review consistently improves accuracy across these benchmarks, although the magnitude of the gains varies by task.

\paragraph{The pattern extends across sender families.}
Both cross-family senders improve the receiver on every math/code task.
Qwen3.8-Max with review has the highest macro-average and leads on all five tasks. The cross-family results show that the gains are not confined to a within-family pairing, but sender-alone scores and uncertainty estimates are needed to compare transfer efficiency reliably.

\begin{table*}[t]
\centering
 \caption{\textbf{Math and code results.} Senders are frozen and prefill-only.
   POND uses one latent transfer; +Review adds periodic prefill updates.
   Acc.: accuracy/pass@1 (\%, higher is better);
   Token: total decoded receiver tokens, including both thinking
   and final-response tokens (thousands per 100 runs, lower is better);
   Speed: end-to-end latency (s/run, lower is better).
   Bold marks the better of POND and +Review for each metric,
   including ties. Improve compares the bold value with the receiver alone:
   accuracy gain in percentage points, or relative reduction in tokens/latency;
   an upward latency arrow indicates an increase. 
   Avg averages the five tasks.}
\label{tab:main}
\resizebox{\textwidth}{!}{%
\begin{tabular}{l c c ccc ccc ccc}
\toprule
\multirow{2}{*}{\textbf{Tasks}} &
\multirow{2}{*}{\textbf{Metrics}} &
\multirow{2}{*}{\makecell{\textbf{Receiver}\\\textbf{alone}}} &
\multicolumn{3}{c}{\textbf{Sender: Qwen3.8-Max}} &
\multicolumn{3}{c}{\textbf{Sender: GLM-5.3}} &
\multicolumn{3}{c}{\textbf{Sender: DeepSeek-V4-Flash}} \\
\cmidrule(lr){4-6}
\cmidrule(lr){7-9}
\cmidrule(lr){10-12}
& & & \method{} & +Review & Improve & \method{} & +Review & Improve & \method{} & +Review & Improve \\
\midrule

\cellcolor{mathbg} & Acc. & 80.0
& 85.4 & \textbf{86.2} & \up{6.2}
& 84.9 & \textbf{85.5} & \up{5.5}
& 84.6 & \textbf{85.1} & \up{5.1} \\
\cellcolor{mathbg} & Token & 982.4
& 836.7 & \textbf{768.9} & \mockreduce{21.7}
& 861.5 & \textbf{801.4} & \mockreduce{18.4}
& 879.2 & \textbf{829.7} & \mockreduce{15.5} \\
\multirow{-3}{*}{\cellcolor{mathbg}\textbf{AIME}} & Speed & 219.4
& \textbf{190.7} & 199.6 & \mockreduce{13.1}
& \textbf{196.8} & 205.1 & \mockreduce{10.3}
& \textbf{202.3} & 209.7 & \mockreduce{7.8} \\
\midrule

\cellcolor{mathbg} & Acc. & 50.0
& 52.9 & \textbf{53.7} & \up{3.7}
& 52.6 & \textbf{53.1} & \up{3.1}
& 52.2 & \textbf{52.8} & \up{2.8} \\
\cellcolor{mathbg} & Token & 1437.8
& 1211.6 & \textbf{1094.3} & \mockreduce{23.9}
& 1248.9 & \textbf{1152.7} & \mockreduce{19.8}
& 1281.4 & \textbf{1197.6} & \mockreduce{16.7} \\
\multirow{-3}{*}{\cellcolor{mathbg}\makecell{\textbf{OlymMath}\\\textbf{Hard}}} & Speed & 311.7
& \textbf{263.8} & 276.4 & \mockreduce{15.4}
& \textbf{271.6} & 282.3 & \mockreduce{12.9}
& \textbf{280.9} & 290.1 & \mockreduce{9.9} \\
\midrule

\cellcolor{codebg} & Acc. & 81.7
& 83.5 & \textbf{84.3} & \up{2.6}
& 83.2 & \textbf{83.9} & \up{2.2}
& 83.0 & \textbf{83.6} & \up{1.9} \\
\cellcolor{codebg} & Token & 94.6
& 86.1 & \textbf{81.9} & \mockreduce{13.4}
& 88.0 & \textbf{84.7} & \mockreduce{10.5}
& 89.5 & \textbf{86.8} & \mockreduce{8.2} \\
\multirow{-3}{*}{\cellcolor{codebg}\textbf{HumanEval+}} & Speed & 25.9
& \textbf{24.1} & 26.0 & \mockreduce{6.9}
& \textbf{24.7} & 26.6 & \mockreduce{4.6}
& \textbf{25.2} & 27.0 & \mockreduce{2.7} \\
\midrule

\cellcolor{codebg} & Acc. & 64.8
& 66.8 & \textbf{67.7} & \up{2.9}
& 66.4 & \textbf{67.42} & \up{2.4}
& 66.1 & \textbf{66.8} & \up{2.0} \\
\cellcolor{codebg} & Token & 57.9
& 52.8 & \textbf{50.1} & \mockreduce{13.5}
& 54.1 & \textbf{52.2} & \mockreduce{9.8}
& 55.0 & \textbf{53.6} & \mockreduce{7.4} \\
\multirow{-3}{*}{\cellcolor{codebg}\textbf{MBPP+}} & Speed & 16.8
& \textbf{16.0} & 17.3 & \mockreduce{4.8}
& \textbf{16.4} & 17.8 & \mockreduce{2.4}
& \textbf{16.6} & 18.1 & \mockreduce{1.2} \\
\midrule

\cellcolor{codebg} & Acc. & 43.2
& 47.1 & \textbf{48.2} & \up{5.0}
& 46.6 & \textbf{47.4} & \up{4.2}
& 45.2 & \textbf{46.1} & \up{4.6} \\
\cellcolor{codebg} & Token & 371.5
& 319.7 & \textbf{289.8} & \mockreduce{22.0}
& 327.6 & \textbf{305.4} & \mockreduce{17.8}
& 338.2 & \textbf{316.7} & \mockreduce{14.8} \\
\multirow{-3}{*}{\cellcolor{codebg}\makecell{\textbf{LiveCode}\\\textbf{Bench}}} & Speed & 88.3
& \textbf{75.6} & 79.1 & \mockreduce{14.4}
& \textbf{78.9} & 82.0 & \mockreduce{10.6}
& \textbf{81.2} & 84.6 & \mockreduce{8.0} \\
\midrule

\cellcolor{avgbg} & Acc. & 63.94
& 67.14 & \textbf{68.02} & \up{4.08}
& 66.74 & \textbf{67.42} & \up{3.48}
& 66.2 & \textbf{66.9} & \up{2.96} \\
\cellcolor{avgbg} & Token & 588.8
& 501.4 & \textbf{457.0} & \mockreduce{22.4}
& 516.0 & \textbf{479.3} & \mockreduce{18.6}
& 528.7 & \textbf{496.9} & \mockreduce{15.6} \\
\multirow{-3}{*}{\cellcolor{avgbg}\textbf{Avg}} & Speed & 132.4
& \textbf{114.0} & 119.7 & \mockreduce{13.9}
& \textbf{117.7} & 122.8 & \mockreduce{11.1}
& \textbf{121.2} & 125.9 & \mockreduce{8.4} \\
\bottomrule
\end{tabular}%
}
\end{table*}

\subsection{Results on Knowledge Benchmarks}
\label{sec:knowledge}

\paragraph{Gains extend beyond the math/code suite.}
Table~\ref{tab:general} reports improvements for all 12 sender--task pairs.
From a receiver average of 84.77\%, single-transfer \method{} reaches 86.44\%, 86.08\%, and 85.75\% with Qwen3.8-Max, GLM-5.3, and DeepSeek-V4-Flash.
Review increases these to 87.18\%, 86.68\%, and 86.35\%, or 1.58--2.41 points above the receiver.
The ordering of senders by macro-average matches that in math/code, but the absolute improvements are smaller.

\paragraph{Review sensitivity varies by task.}
Review adds 1.01--1.52 points on GPQA-Diamond, the longest-output knowledge task, compared with 0.35--0.62 points on the other three tasks.
This pattern motivates evaluating review's benefit together with its latency cost rather than assuming that repeated sender involvement is equally useful across tasks.

\begin{table*}[t]
\centering
\caption{\textbf{Knowledge benchmark results.}
Configurations, metrics, and metric-wise bolding follow Table~\ref{tab:main}.
Improve reports accuracy gains (percentage points) or relative token/latency reductions (\%) against the receiver; upward latency arrows denote increases.
Avg averages all four tasks.
Differences and reductions are computed before rounding.}
\label{tab:general}
\resizebox{\textwidth}{!}{%
\begin{tabular}{l c c ccc ccc ccc}
\toprule
\multirow{2}{*}{\textbf{Tasks}} &
\multirow{2}{*}{\textbf{Metrics}} &
\multirow{2}{*}{\makecell{\textbf{Receiver}\\\textbf{alone}}} &
\multicolumn{3}{c}{\textbf{Sender: Qwen3.8-Max}} &
\multicolumn{3}{c}{\textbf{Sender: GLM-5.3}} &
\multicolumn{3}{c}{\textbf{Sender: DeepSeek-V4-Flash}} \\
\cmidrule(lr){4-6}
\cmidrule(lr){7-9}
\cmidrule(lr){10-12}
& & &
\method{} & +Review & Improve &
\method{} & +Review & Improve &
\method{} & +Review & Improve \\
\midrule

\cellcolor{genbg} & Acc. & 76.77
& 79.29 & \textbf{80.81} & \up{4.04}
& 78.79 & \textbf{79.80} & \up{3.03}
& 78.28 & \textbf{79.29} & \up{2.52} \\
\cellcolor{genbg} & Token & 660.3
& 568.9 & \textbf{504.7} & \mockreduce{23.6}
& 585.1 & \textbf{529.8} & \mockreduce{19.8}
& 601.4 & \textbf{548.6} & \mockreduce{16.9} \\
\multirow{-3}{*}{\cellcolor{genbg}\makecell{\textbf{GPQA}\\\textbf{Diamond}}}
& Speed & 149.2
& \textbf{128.6} & 136.9 & \mockreduce{13.8}
& \textbf{134.8} & 142.1 & \mockreduce{9.7}
& \textbf{139.7} & 146.5 & \mockreduce{6.4} \\
\midrule

\cellcolor{genbg} & Acc. & 89.58
& 90.47 & \textbf{90.86} & \up{1.28}
& 90.22 & \textbf{90.63} & \up{1.05}
& 90.04 & \textbf{90.39} & \up{0.81} \\
\cellcolor{genbg} & Token & 102.4
& 95.6 & \textbf{89.7} & \mockreduce{12.4}
& 97.1 & \textbf{92.4} & \mockreduce{9.8}
& 98.5 & \textbf{94.9} & \mockreduce{7.3} \\
\multirow{-3}{*}{\cellcolor{genbg}\makecell{\textbf{MMLU}\\\textbf{Redux}}}
& Speed & 27.1
& \textbf{25.7} & 28.1 & \mockreduce{5.2}
& \textbf{26.3} & 28.8 & \mockreduce{3.0}
& \textbf{26.8} & 29.4 & \mockreduce{1.1} \\
\midrule

\cellcolor{genbg} & Acc. & 78.62
& 80.93 & \textbf{81.47} & \up{2.85}
& 80.51 & \textbf{81.08} & \up{2.46}
& 80.12 & \textbf{80.74} & \up{2.12} \\
\cellcolor{genbg} & Token & 286.7
& 246.2 & \textbf{218.5} & \mockreduce{23.8}
& 253.8 & \textbf{229.7} & \mockreduce{19.9}
& 262.1 & \textbf{239.6} & \mockreduce{16.4} \\
\multirow{-3}{*}{\cellcolor{genbg}\makecell{\textbf{MMLU}\\\textbf{Pro}}}
& Speed & 69.4
& \textbf{60.5} & 64.3 & \mockreduce{12.8}
& \textbf{63.4} & 67.1 & \mockreduce{8.6}
& \textbf{65.7} & 69.2 & \mockreduce{5.3} \\
\midrule

\cellcolor{genbg} & Acc. & 94.11
& 95.05 & \textbf{95.56} & \up{1.45}
& 94.80 & \textbf{95.22} & \up{1.11}
& 94.54 & \textbf{94.97} & \up{0.86} \\
\cellcolor{genbg} & Token & 145.8
& 127.9 & \textbf{116.3} & \mockreduce{20.2}
& 131.6 & \textbf{121.5} & \mockreduce{16.7}
& 135.2 & \textbf{126.8} & \mockreduce{13.0} \\
\multirow{-3}{*}{\cellcolor{genbg}\makecell{\textbf{ARC}\\\textbf{Challenge}}}
& Speed & 37.6
& \textbf{33.1} & 35.6 & \mockreduce{12.0}
& \textbf{34.5} & 37.0 & \mockreduce{8.2}
& \textbf{35.7} & 38.4 & \mockreduce{5.1} \\
\midrule

\cellcolor{avgbg} & Acc. & 84.77
& 86.44 & \textbf{87.18} & \up{2.41}
& 86.08 & \textbf{86.68} & \up{1.91}
& 85.75 & \textbf{86.35} & \up{1.58} \\
\cellcolor{avgbg} & Token & 298.8
& 259.7 & \textbf{232.3} & \mockreduce{22.3}
& 266.9 & \textbf{243.4} & \mockreduce{18.6}
& 274.3 & \textbf{252.5} & \mockreduce{15.5} \\
\multirow{-3}{*}{\cellcolor{avgbg}\textbf{Avg}}
& Speed & 70.8
& \textbf{62.0} & 66.2 & \mockreduce{12.5}
& \textbf{64.8} & 68.8 & \mockreduce{8.6}
& \textbf{67.0} & 70.9 & \mockreduce{5.4} \\
\bottomrule
\end{tabular}%
}
\end{table*}

\subsection{Accuracy--Latency Trade-offs and Periodic Review}
\label{sec:efficiency}

\paragraph{Shorter outputs do not imply proportional latency savings.}
Relative to the receiver, +Review reduces average output volume by 15.6--22.4\% on math/code and 15.5--22.3\% on knowledge tasks (Tables~\ref{tab:main} and~\ref{tab:general}).
The corresponding latency reductions are only 4.9--9.6\% on math/code and range from a 6.5\% reduction to a 0.1\% increase on knowledge tasks.
Sender prefill and latent conditioning still incur computation, so fewer receiver tokens do not establish proportional savings in total compute.

\paragraph{Review improves quality more consistently than latency.}
Review improves accuracy and reduces output volume relative to single-transfer \method{} in all 27 sender--task pairs.
Its incremental macro-average gain is 0.7--0.9 points on math/code and 0.60--0.74 points on knowledge tasks.
However, the additional prefill passes increase latency relative to single transfer in every pair.
Review remains faster than the receiver on the longest-output tasks, including AIME, OlymMath-Hard, LiveCodeBench, and GPQA-Diamond, but exceeds receiver latency for every sender on the shorter HumanEval+, MBPP+, and MMLU-Redux tasks.
This pattern is consistent with review overhead being easier to amortize over longer generations, although component-level timing is needed to establish the mechanism.

The benefit of review is consistent with a trajectory-dependent effect. Its incremental gains are largest on long-output tasks such as AIME, LiveCodeBench, and GPQA-Diamond, where guidance computed from the initial prompt may become less informative as the response develops. On shorter tasks, review provides smaller accuracy gains while adding proportionally greater latency. This pattern suggests that review is most useful when the evolving reasoning trajectory contains information not available in the initial prompt.

\begin{table}[t]
\centering
\small
\caption{\textbf{Training and review controls.}
Suite-average accuracy. Receiver-only training uses math/code data; \method{} uses Qwen3.8-Max as the sender.}
\label{tab:receiver_training_control}
\setlength{\tabcolsep}{8pt}
\begin{tabular}{lcc}
\toprule
Method & Math/code & Knowledge \\
\midrule
Frozen receiver       & 63.94 & 84.77 \\
Receiver-only training & 64.2 & 83.45 \\
\method{}              & 66.9 & 86.44 \\
\method{} + Review     & \textbf{68.02} & \textbf{87.18} \\
\bottomrule
\end{tabular}
\end{table}

\paragraph{Aggregate improvements are not uniform per-task speedups.}
All three review configurations improve mean accuracy and latency over the receiver on math/code. On knowledge tasks, Qwen3.8-Max and GLM-5.3 improve both means, while DeepSeek-V4-Flash improves accuracy but is 0.1 seconds slower on average.
Because longer tasks contribute more to the ratio-of-means latency summary, suite-level improvements do not imply a speedup for every workload or justify a fixed review policy under every latency constraint.
Qwen3.8-Max has the best aggregate accuracy, output volume, and latency in both suites, although these results do not establish a general ranking of sender quality.

\paragraph{Receiver-only training does not explain the full gains.} As shown in Table~\ref{tab:receiver_training_control}, training the receiver directly on math/code raises its suite average from 63.94\% to 64.2\%, but lowers knowledge accuracy from 84.77\% to 83.45\%, suggesting a small loss in general-knowledge performance. Jointly training the receiver and bridge with latent guidance improves both suites, reaching 66.9\% on math/code and 86.44\% on knowledge with a single transfer, with further gains from review.

\begin{figure}[!ht]
    \centering
    \includegraphics[width=0.95\linewidth]{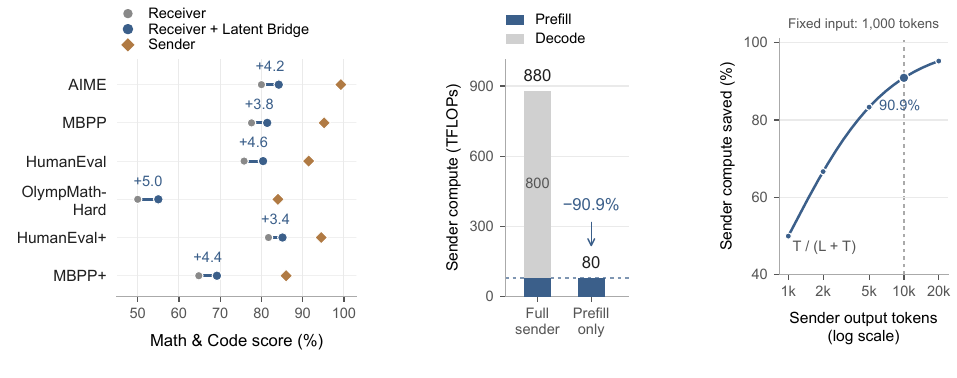}
    \caption{\textbf{Accuracy gains and sender compute savings.}
\textbf{Left:} Math and code scores for the receiver (Qwen3.5-4B), the receiver with \method, and the sender (GLM-5.3); labels give the gain in points.
\textbf{Center:} Estimated sender FLOPs for GLM-5.3 (40B active) with 1k input and 10k output tokens. Prefill-only guidance reduces sender compute from 880 to 80 TFLOPs.
\textbf{Right:} Fraction of sender compute saved, $T/(L+T)$, as a function of output length $T$ with $L=1$k.
Estimates use $2P_{\mathrm{active}}$ FLOPs per token and exclude attention, routing, communication, and bridge or receiver inference. The panels are not a matched end-to-end comparison of accuracy and cost.}
    \label{fig:compute}
\end{figure}

\subsection{Accuracy Gains and Sender Compute}
\label{sec:sender_compute}

Figure~\ref{fig:compute} relates the accuracy gained from latent guidance to the sender-side computation avoided relative to textual advising. The panels provide complementary views of quality and analytical compute rather than matched end-to-end operating points. We highlight three observations.

\paragraph{A single transfer recovers part of the sender advantage.}
Across the benchmarks in the left panel, \method consistently improves the receiver, but closes different fractions of the receiver--sender gap. The recovered fraction is smaller on tasks where the sender's advantage is larger. This suggests that prompt representations convey useful task information without reproducing all of the computation performed during sender-side generation.

\paragraph{Review helps most on longer trajectories.}
Periodic review yields its largest gains on long-output tasks (AIME,
LiveCodeBench, GPQA-Diamond) and smaller gains on shorter ones. One explanation
is that guidance derived from the initial prompt grows stale as the response
unfolds, whereas review conditions on the intermediate trajectory. Aggregate
results, however, cannot separate better planning from error correction or
reduced verbosity.

\paragraph{Sender-side savings scale with the avoided response length.}
Under a parameter-multiply approximation, sender generation costs
\(2P_{\mathrm{active}}(L+T)\) versus \(2P_{\mathrm{active}}L\) for a single
prefill, saving a fraction \(T/(L+T)\). With cached delta-only review, this
becomes \((T-\sum_r \Delta U_r)/(L+T)\), positive whenever
\(\sum_r \Delta U_r < T\). These estimates exclude attention, communication,
and receiver inference, as well as review overheads, and thus reflect
sender-side savings rather than end-to-end speedup.

\section{Conclusion}
\label{sec:conclusion}

We introduced \method, which decouples frontier-model prefill from small-model
decoding through a learned latent interface. A trainable bridge converts a
frozen sender's final-layer representations into latent KV entries, from which
a receiver generates the full answer without any textual handoff. Periodic
review extends this to partial responses, refreshing guidance while preserving
the receiver's generation state. The central contribution is thus an interface
through which a large model participates in generation without decoding text
itself. Establishing its advantage over other collaboration strategies will
require matched-compute text baselines, controlled probes of sender
information, and uncertainty estimates; adaptive review policies and bounded
latent memory are natural next steps.

\subsection*{AI use statement}

We used generative AI tools to assist with drafting and editing the manuscript, general code editing, and preparing figures and charts. We take responsibility for the final content of this work, including all text, code, and figures produced with AI assistance.

\appendix
\newpage
\section{Appendix}
\label{app:training}

This appendix documents the implementation and training semantics used in
the experiments reported in Section~\ref{sec:experiments}. The completed system
used final-layer sender features, a frozen sender, joint receiver/bridge
training, dense latent attention, and cached delta-only review. All numerical
hyperparameters were tied to the corresponding run manifests, and all
benchmark measurements reported in the paper were produced by these runs.

\subsection{Objective and gradient paths}
\label{app:objective}
Equation~\ref{eq:training} averaged response-token negative
log-likelihood over valid targets. Prompt and padding positions received
zero loss. Sender features were detached, while receiver operations remained
differentiable with respect to both receiver weights and injected KV entries.
The trained bridge consisted of the two-layer compressor, per-layer K/V
projections, key RMS-normalization scales, and scalar gates. The dense
configuration used in the main experiments had no indexer parameters.

For the bridge-only ablation, we kept the receiver weights fixed while
retaining autograd through receiver operations to the bridge. For the
no-message jointly trained receiver ablation, we optimized the receiver
without latent messages. We assigned distinct condition names and artifact
hashes to the original receiver, the receiver-only training baseline, the
bridge-only ablation, and the jointly trained receiver/bridge system.

\paragraph{Teacher-forced review.}
For each fixed tokenized input/answer sequence, the causal sender computed all
features in one full pass. Review $r$ selected only the features associated
with its newly observed segment. The training attention mask enforced both
span alignment and review release, preventing review features from becoming
visible to earlier receiver queries. Consequently, the receiver never had
access to future-answer information during training. Single-pass feature
extraction reduced training cost without changing the causal message mask.

\paragraph{Train--inference equivalence.}
We validated full causal sender execution against cached chunk execution on
identical token sequences and boundary schedules. The comparison used matched
positions, tokenization, architecture, and numerical settings. We likewise
validated receiver full-pass execution with per-query masks against
incremental execution with preserved historical states. The two paths agreed
within the numerical tolerance defined in the run manifest. This validation
does not imply bitwise equality under arbitrary kernels: dropout,
implementation-dependent floating-point order, and stateful kernels can alter
the result. Moreover, teacher forcing uses reference prefixes whereas
deployment uses sampled prefixes, so the validation does not remove the
resulting distribution shift.

\section{Position Alignment and Message Availability}
\label{app:alignment}

We treated the position identifier and release threshold as separate
quantities. The former controlled rotary phase, whereas the latter prevented
a receiver query from reading an unavailable message. For a shared serialized
text, Equation~\ref{eq:alignment} assigned each sender token to the first receiver
token covering its complete source span. Multiple sender tokens could map to
one receiver position; they remained distinct KV entries and did not consume
or shift text-cache positions.

\paragraph{Offset convention.}
We represented offsets in a common UTF-8 byte coordinate system and converted
tokenizer-provided offsets before alignment. Sender and receiver offsets were
computed from the same normalized and serialized string, including recorded
chat delimiters and special tokens. We excluded non-text tokens without an
unambiguous receiver counterpart from latent memory rather than assigning
guessed spans. Inputs whose normalization or serialization could not be
reconciled were marked as interface failures and excluded according to the
predefined evaluation rule.

\paragraph{Incremental tokenization.}
At every review, the implementation verified that the new sender token
sequence extended the cached token prefix. We used a prefix-stable streaming
tokenization policy and retained token IDs and source spans at every boundary.
When a cache fallback was required and the existing tokens and positions were
unchanged, the implementation recomputed the sender prefix, charged the full
recomputation, and appended only newly available states. If old tokens or
source spans changed, the implementation recorded a tokenization failure and
terminated the request rather than silently altering receiver history. Full
request restarts, where applicable, were recorded separately together with
their work and generated output.

\paragraph{Causal visibility claim.}
Assume (i) sender state $j$ depends only on its fixed serialized source
prefix; (ii) alignment $p_j$ covers that complete prefix span; (iii)
release decisions use only already observed text and runtime state; and
(iv) the receiver is causal and never recomputes historical states with later
messages. Then prediction of $y_t$ under Equation~\ref{eq:visibility} cannot
depend on $y_t$ or later response tokens. Indeed, its query position is
$q=n_x+t-1$; every visible latent satisfies $p_j\leq q$ and $\rho_r\leq q$,
so its source and release are both in the observed prefix. Text attention and
recurrent states are also causal. Induction through receiver layers and
decoding steps proves the claim. This is a conditional correctness property,
not an accuracy theorem.

\paragraph{Next-token indexing.}
The receiver query at $q=n_x+t-1$ predicted $y_t$. A review issued after
sampling $y_b$ had release position $\rho=n_x+b$ and first affected prediction
of $y_{b+1}$. For each latent entry $j$, the implementation masked the entry
unless $q\geq\max(p_j,\rho)$. This rule also applied when its rotary position
was early in the prompt. Padding was never visible. Receiver prompt queries
used the same positional causal mask and therefore did not attend to the full
prompt message prematurely.

\section{Gated Attention and Sparse Reading}
\label{app:attention}
\label{app:sparse}

For one query/head, let $C_{\mathrm{txt}}=\sum_i\exp(a_i^{\mathrm{txt}})$
and $C_z=\sum_j\exp(a_j^z)$ sum valid scaled logits. With separately
normalized outputs $A^{\mathrm{txt}}$ and $A^z$,
\begin{equation}
 A^{\mathrm{joint}}=(1-\alpha)A^{\mathrm{txt}}+\alpha A^z,
 \qquad \alpha=\frac{C_z}{C_{\mathrm{txt}}+C_z}.
 \label{eq:app_attention_decomposition}
\end{equation}
This proves Equation~\ref{eq:residual_view} before dropout. Our implementation
combined log-sum-exp statistics rather than exponentiating large raw logits.
When latent memory was empty, it returned text attention. Duplicate messages
were not treated as neutral because adding entries changes the normalizer.

\paragraph{Zero-gate initialization.}
At $g_\ell=0$, the local attention equals the original receiver output.
The gate derivative is
\begin{equation}
 \frac{\partial A_\ell}{\partial g_\ell}
 =(1-\tanh^2g_\ell)(A_\ell^{\mathrm{joint}}-A_\ell^{\mathrm{txt}}).
\end{equation}
The latent K/V gradient was initially multiplied by zero, while the gate
could receive a nonzero gradient. With all gates initialized to zero, the
compressor received no likelihood gradient on the first step in the absence
of another loss path. Receiver weights continued to receive ordinary
language-model gradients under joint training. Subsequent optimization was
allowed to move the receiver away from the original checkpoint.

\paragraph{Sparse variant.}
The dense configuration retained every visible entry and used no sequence
compression ($m=1$). In the completed sparse evaluation, top-$k$ reading
applied Equation~\ref{eq:topk} per query head after masking, with stable
entry-index tie breaking. When fewer than $k$ entries were visible, the
implementation retained all of them; when none were visible, it returned text
attention. The latent weighted sum and its log-normalizer used only retained
entries, while text entries were never pruned. Exact query--key scoring still
touched every visible latent key, so top-$k$ reduced selected-value
aggregation rather than making total runtime independent of memory size. We
trained the reference model densely and evaluated sparse deployment in a
separate quality/cost sweep.

\paragraph{Truncation error at a fixed query.}
Suppose the text and latent values have norm at most $B$. Let $\delta$ be the
probability mass that the \emph{full joint softmax} assigns to latent entries
omitted by sparse reading. Denote the dense and sparse gated outputs by $A$
and $\widetilde A$, using the same gate and text branch. Conditioning the full
softmax on retained entries gives
$A^{\mathrm{joint}}=(1-\delta)\widetilde A^{\mathrm{joint}}+
\delta A^{\mathrm{omit}}$. Therefore
\begin{equation}
 \|A-\widetilde A\|\leq 2B|\tanh g|\,\delta.
 \label{eq:sparse_bound}
\end{equation}
Both conditional weighted sums have norm at most $B$, which establishes the
inequality by the triangle inequality. If nothing is omitted, the outputs
agree. A small $k$ alone does not bound $\delta$, so sparse reading has no
unconditional approximation guarantee. This bound is local to one attention
output; later layers and autoregressive sampling may amplify a perturbation.
It does not bound answer accuracy, and estimating the exact omitted mass
itself requires full scoring.

\section{Inference and Cache Semantics}
\label{app:review}
\label{app:cached_inference}

Algorithm~\ref{alg:inference} specifies the incremental-prefill inference
procedure used in our evaluation. Here $B_\phi(x)$ abbreviates compression of
sender features, and $\mathcal{M}$ denotes the collection of per-layer
memories. Sender input segments were serialized using the procedure in
Appendix~\ref{app:alignment}.

\par\medskip
\refstepcounter{algorithm}\label{alg:inference}
\noindent\begin{minipage}{\linewidth}
\hrule\vspace{0.5em}
\small
\textbf{Algorithm \thealgorithm: Inference with \method}\par
\emph{Inputs:} $x$; review boundaries $b_1<\cdots<b_R$; output limit
$N_{\max}$.
\begin{enumerate}[leftmargin=1.6em,itemsep=2pt,topsep=4pt]
\item Set $b_0\gets0$, $r\gets1$, and
$n_x\gets |x|_{\mathrm{receiver}}$. Serialize $x$ and record sender token IDs,
spans, and aligned positions.
\item Prefill the sender on $x$ while retaining its cache; build
$\mathcal{M}^{(0)}$ from $Z^{(0)}=B_\phi(x)$. Set
$\mathcal{M}\gets\mathcal{M}^{(0)}$ and prefill the receiver on $x$ using
Equation~\ref{eq:visibility}.
\item For $t=1,\ldots,N_{\max}$, sample $y_t$. Stop if it is \texttt{EOS} or
$t=N_{\max}$.
\item If $r\leq R$ and $t=b_r$, serialize the available text and verify that
sender prefix IDs and spans are unchanged. Terminate and record an interface
failure if prefix stability or position alignment fails. Otherwise prefill the
sender on $y_{b_{r-1}+1:b_r}$ using its retained cache, build $Z^{(r)}$, append
its KV entries to $\mathcal{M}$ using Equation~\ref{eq:memory_update}, set
$\rho_r=n_x+t$, and increment $r$.
\item Process $y_t$ in the receiver at position $n_x+t$ with $\mathcal{M}$
using Equation~\ref{eq:residual_view} and, optionally,
Equation~\ref{eq:topk}; then continue from Step~3.
\end{enumerate}
\vspace{0.2em}\hrule
\end{minipage}
\par\medskip

\paragraph{Historical states were not rewritten.}
Each cached receiver state recorded the messages available when it was
computed. Reviews updated future queries without changing historical states.
Thus, the implementation was causal by induction on decoding steps but was
not equivalent to recomputing the entire prefix with the newest memory visible
at all positions. Hybrid receiver state, including non-attention recurrent
state, was preserved together with the text KV cache. Reviews did not retract
tokens that had already been emitted.

\paragraph{Delta-only versus full-message review.}
The evaluated implementation appended each new sender token once.
Re-prefilling and appending the entire context at every boundary would
duplicate earlier latent entries and alter attention normalization and
storage; we did not use that variant in Algorithm~\ref{alg:inference}. When a
cache fallback recomputed old sender states, it appended only new entries and
recorded full-prefix sender work separately from delta-only receiver storage.

\section{Resource Accounting}
\label{app:cost}
\subsection{Sender work}
\label{app:sender_cost}
Let $P_s$ be sender active parameters, $L_s$ its prompt-token count,
$D_s$ the generated-token count of a hypothetical sender-decoding baseline,
and $\Delta U_r$ the newly processed sender tokens in review $r$. Under a
parameter-multiply-only approximation,
\begin{align}
 F_{\mathrm{sender,decode}}&\approx2P_s(L_s+D_s),\\
 F_{\mathrm{sender,initial}}&\approx2P_sL_s,\\
 F_{\mathrm{sender,cached}}&\approx2P_s\left(L_s+\sum_r\Delta U_r\right).
 \label{eq:app_cached_cost}
\end{align}
The corresponding avoided fraction is
\begin{equation}
 \eta_{\mathrm{cached}}\approx
 \frac{D_s-\sum_r\Delta U_r}{L_s+D_s}.
 \label{eq:app_review_saving}
\end{equation}
Savings are positive only when reviewed sender tokens are fewer than the
avoided sender output tokens. This comparison is against sender decoding, not
receiver-alone inference. For $P_s=40$ billion, $L_s=1000$, $D_s=10000$, and
no reviews, the analytical approximation gives 80 versus 880 TFLOPs, or
90.9\% avoided parameter-multiply work. We treated these as analytical
estimates rather than GPU measurements.

For recorded no-cache fallbacks, a full sender context of length $U_r$ was
reprocessed at each review, costing $2P_s(L_s+\sum_rU_r)$. Under common-token
boundaries $b_r=rq$, $\sum_rU_r\approx RL_s+qR(R+1)/2$. We applied this
quadratic full-prefill expression only to no-cache fallbacks. Cached review
still attended to the retained sender prefix and preserved sender state; we
therefore did not equate token-pass counts with wall-clock time across chunk
sizes, attention kernels, or devices.

\subsection{Latent memory, communication, and attention}
\label{app:memory}
Let $V_0$ be the number of prompt sender tokens retained as valid latent
entries and $\Delta V_r$ the retained entries in review $r$. Unlike the
processed lengths $L_s$ and $\Delta U_r$, these excluded unmapped special
tokens. Delta-only storage was
$M_z=V_0+\sum_r\Delta V_r\leq L_s+\sum_r\Delta U_r$; equality held only when
every processed token was retained. Sender FLOPs included all processed
tokens, including excluded special tokens. With $L_a$ adapted layers and
$\beta_{\mathrm{KV}}$ bytes per scalar, added receiver KV storage was
\begin{equation}
 B_{\mathrm{latent\ KV}}=2\beta_{\mathrm{KV}}L_aM_z n_{\mathrm{kv}}d_h.
 \label{eq:app_kv_bytes}
\end{equation}
This quantity excluded sender caches, text caches, weights, and temporary
activations. For nonuniform KV layouts, we summed the corresponding quantity
over layers. Sharing $Z$ reduced one-way transmission to
$\beta_zM_zd_z$ bytes when compression ran sender-side and KV projection ran
receiver-side; it did not remove the receiver's per-layer cache. Review also
transmitted text-prefix increments in the reverse direction.

For $n_q$ receiver query heads and $M_z(t)$ visible latent entries, dense
response attention added approximately
\begin{equation}
 F_{\mathrm{latent\ attention}}\approx
 4L_a n_qd_h\sum_{t=1}^{N}M_z(t),
 \label{eq:app_attention_cost}
\end{equation}
excluding softmax, projections, prompt attention, and duplicated text work.
KV storage scaled with $n_{\mathrm{kv}}$, not $n_q$. With exact top-$k$
selection, scoring remained linear in $M_z(t)$ while only selected-value
aggregation scaled with $k$. More frequent reviews could increase the sum even
at a fixed final memory length because entries became visible earlier.

\paragraph{End-to-end accounting.}
For the serial implementation, we decomposed latency as
\begin{equation}
 \tau=\tau_s+\tau_{\mathrm{bridge}}+\tau_{\mathrm{transfer}}
      +\tau_{r,\mathrm{prefill}}+\tau_{r,\mathrm{decode}}.
\end{equation}
The sender term included every review and cache fallback. Where stages
overlapped, we measured the critical path instead of adding overlapping
intervals. We assessed speedup against receiver-alone inference only when the
measured reduction in receiver work exceeded all added overheads. We did not
treat shorter output length alone as a total-FLOPs or latency measurement.

\section{Reproducibility and Scope}
\label{app:reproducibility}
\label{app:limitations}

\paragraph{Configuration status.}
The completed run manifests recorded model and tokenizer revisions,
hidden-state access, bridge dimensions, adapted layers, normalization, gate
initialization, training data and splits, optimizer and schedule, precision,
random seeds, checkpoint selection, review policy, and decoding settings.
All aggregate tables were generated from per-run artifacts associated with
these manifests rather than reconstructed from summary scores. Sender models
were accessed through interfaces exposing the required hidden states; no
text-only sender API was used for the latent interface.

\subsection{Experimental Configuration}
\label{app:pilot}
We used $d_b=2048$, $d_z=512$, key RMSNorm epsilon $10^{-6}$,
dense attention without dropout, and adapters in all full-attention receiver
layers. We optimized with AdamW using receiver and bridge learning rates of
$5\times10^{-6}$ and $10^{-4}$, betas $(0.9,0.95)$, weight decay $0.01$,
gradient clipping at 1, an effective batch size of 32 examples, one epoch, and
3\% warm-up followed by cosine decay. Norms and loss used FP32, and model
computation used BF16.

We trained with reference-prefix reviews every 512 receiver tokens. At the same
checkpoint, we evaluated no-review inference and review strides of 256, 512,
and 1024 tokens. Evaluation used greedy generation with one output per problem.
The final response was capped at 4,096 tokens, within an 8,192-token limit on
the combined prompt and final response. Both limits exclude the model's
thinking segment, so the token counts in Tables~\ref{tab:main}
and~\ref{tab:general}, which include thinking tokens, can exceed them.
Oversize examples were identified and rejected using the predefined rule rather
than silently truncated. We completed independent training runs with seeds 11,
23, and 47. On the densely trained model, we also completed the sparse sweep
with $k \in \{32, 64, 128\}$ and $m=1$.

The run manifests record the exact model and tokenizer revisions, training
and evaluation dataset manifests, licenses, software environment, and
hardware configuration. Pre-run validation confirmed hidden-state access and
agreement between the implementation and the equations; completion records
and checkpoint hashes were stored for every successful training run.

\paragraph{Evaluation record.}
For every benchmark, we recorded the version, sample count, prompt template,
reasoning mode, answer extraction procedure, executable-test settings and
timeouts, truncation and failure handling, and decontamination procedure. We
archived per-example predictions, correctness, latency, output length, review
count, sender-token count, cache-fallback count, and peak memory. Timing
records included hardware, software, parallelism, concurrency, warm-up,
synchronization, repetition count, and whether loading, queueing, and
communication were included.

\paragraph{Attribution and uncertainty.}
The completed comparison included the original receiver, a data- and
budget-matched receiver-only training baseline, and the jointly trained
receiver/bridge system. We additionally evaluated the frozen-receiver bridge
ablation, constant and shuffled messages, and matched-budget text advising as
separate controls. Review-frequency comparisons used the same
receiver/bridge checkpoint. Paired uncertainty estimates resampled the same
problems across systems and grouped multiple samples from each problem.
Training-seed variation used the three completed independent runs, and timing
variation was estimated from repeated synchronized measurements.

\paragraph{Limitations.}
Hidden-state access limits black-box deployment. Incremental review requires
prefix-stable tokenization, available sender state, and custom receiver
kernels. Compressed states are not automatically private or safe to transmit.
Review affects only future predictions and cannot certify or repair past
outputs. The reported downstream accuracy and efficiency remain specific to
the evaluated models, datasets, configurations, and comparison baselines;
they do not establish a universal advantage over all collaboration methods.

\bibliography{iclr2027_conference}

\end{document}